\documentclass[10pt,twocolumn,letterpaper]{article}

\usepackage{iccv}
\usepackage{times}
\usepackage{epsfig}
\usepackage{graphicx}
\usepackage{amsmath}
\usepackage{amssymb}
\usepackage{epsfig}
\usepackage{graphicx}
\usepackage{amsmath}
\usepackage{amssymb}
\usepackage{booktabs}

\usepackage[pagebackref=true,breaklinks=true,letterpaper=true,colorlinks,bookmarks=false]{hyperref}

 \iccvfinalcopy 

\ificcvfinal\fi

\begin{document}

\title{Unsupervised Discovery of Actions in Instructional Videos}


\author{AJ Piergiovanni\\
Robotics at Google\\
\and
Anelia Angelova \\
Robotics at Google\\
\and
Michael Ryoo\\
Robotics at Google\\
\and
Irfan Essa\\
Google Research\\
}

\maketitle

\begin{abstract}

In this paper we address the problem of automatically discovering atomic actions in unsupervised manner from instructional videos. Instructional videos contain complex activities and are a rich source of information for intelligent agents, such as, autonomous robots or virtual assistants, which can, for example, automatically `read' the steps from an instructional video and execute them. However, videos are rarely annotated with atomic activities, their boundaries or duration.
We present an unsupervised approach to learn atomic actions of structured human tasks from a variety of instructional videos. We propose a sequential stochastic autoregressive model for temporal segmentation of videos, which 
   learns to represent and discover the sequential relationship between different atomic actions of the task, and which provides automatic and unsupervised self-labeling for videos. Our approach outperforms the state-of-the-art unsupervised methods with large margins. We will open source the code.  
 
\end{abstract}

\section{Introduction}
Instructional videos cover a wide range of tasks: cooking, furniture assembly, repairs, etc. The availability of online instructional videos for almost any task provides a valuable resource for learning, especially in the case of learning robotic tasks. 
So far, the primary focus of activity recognition has been on supervised classification or detection of discrete actions in videos, such as sports actions~\cite{THUMOS14,yeung2015every,UCF101} or in-home activities, e.g.~\cite{sigurdsson2016hollywood,das2019toyota} using fully annotated videos. 
However, instructional videos are rarely annotated with atomic action-level instructions.
Several works have studied weakly-supervised settings where the order or presence of actions per-video is given, but not their duration \cite{richard2017weakly,huang2016connectionist}. In this work, we propose a method to learn to segment instructional videos in atomic actions in an unsupervised way, i.e., without any annotations. To do this, we take advantage of the structure in instructional videos: they comprise complex actions which inherently consist of smaller atomic actions with predictable order. 
While the temporal structure of activities in instructional videos is strong, there is high variability of the visual appearance of actions, which makes the task, especially in its unsupervised setting, very challenging. For example, videos of preparing a salad can be taken in very different environments, using kitchenware and ingredients of varying appearance.



\begin{figure}
\includegraphics[width=\linewidth]{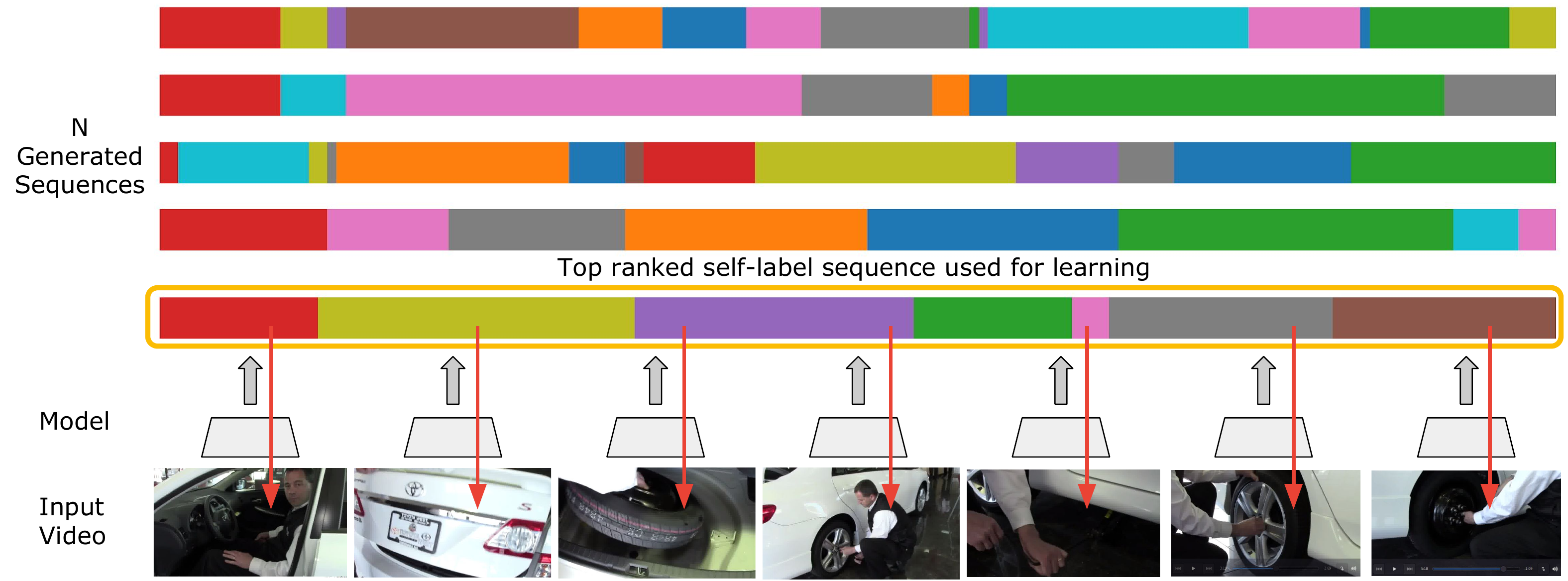}
\caption{Overview: Our model generates multiple sequences for each video which are ranked based on several constraints (colors represent different actions). The top ranked sequence is used as self-labels to train the action segmentation model. This processes is repeated until convergence. No annotations are used.}
    \label{fig:motivation}
\end{figure}

The central idea is to learn a stochastic model that generates multiple, different candidate sequences, which can be ranked based on instructional video constraints. The top ranked sequence is used as self-labels to train the action segmentation model. By iterating this process in an EM-like procedure, the model converges to a good segmentation of actions (Figure~\ref{fig:motivation}). In contrast to previous weakly~\cite{richard2017weakly,huang2016connectionist} and unsupervised~\cite{alayrac2016unsupervised,kukleva2019unsupervised} action learning works, our method only requires input videos, no further text, actions, or other annotations are used.



\begin{figure*}[t]
    \centering
    \includegraphics[width=0.79\linewidth]{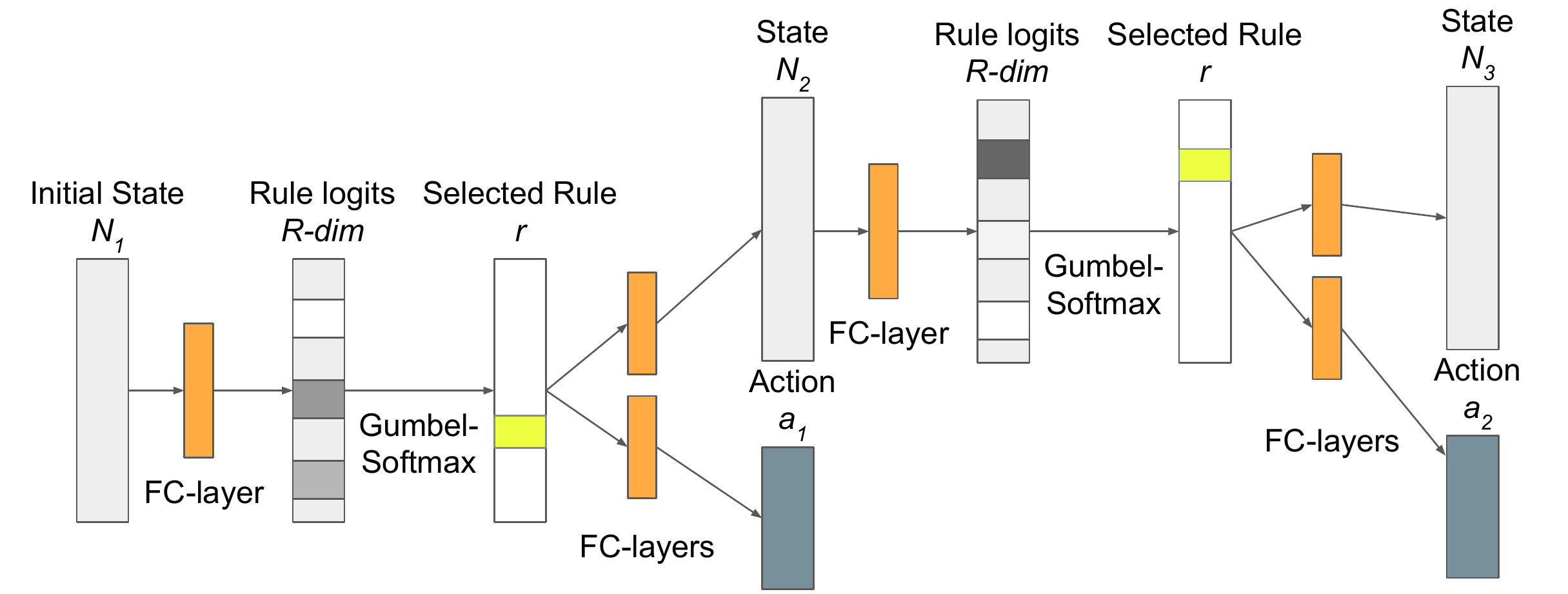}
    \caption{
    Overview of the stochastic recurrent model which generates an output action per step and a latent state (which will in turn generate next actions). Each time the model is run, a different rule is selected, thanks to the Gumbel-Softmax trick, leading to a different action and state. This results in multiple 
    sequences (see text for more details).
    }
    \label{fig:grammar-overview}
\vspace{-0.2cm}
\end{figure*}


We evaluate the approach on multiple datasets and compare to previous methods on unsupervised action segmentation. We also compare to weakly-supervised and supervised baselines. Our unsupervised method outperforms all state-of-the-art models, in some cases considerably, with performance at times outperforming weakly-supervised methods. 

Our contributions are (1) a stochastic model capable of capturing multiple possible sequences, (2) a set of constraints and training method that is able to learn to segment actions without any labeled data.

\section{Related Work}
Studying instructional videos has gained a lot of interest recently \cite{alayrac2016unsupervised,miech2019howto100m,coin,procel}, 
largely fueled by advancements in feature learning and  
activity recognition for videos~\cite{carreira2017quo,xie2018rethinking,tran2018closer,feichtenhofer2018slowfast,Ryoo2020AssembleNet}. 
However, most work on activity segmentation has focused on the fully-supervised case \cite{shou2017cdc,zhu2017tornado}, which requires per-frame labels of the occurring activities.

Since it is expensive to fully annotate videos, weakly-supervised activity segmentation has been proposed. Initial works use movie scripts to obtain weak estimates of actions \cite{laptev2008learning,marszalek2009actions} or  localize actions based on related web images \cite{gan2016webly,gan2016you,sun2015temporal}. \cite{bojanowski2014weakly} perform weakly-supervised segmentation when assuming the ordering was given, both during training and test time. Temporal ordering constraints \cite{huang2016connectionist} or language \cite{anne2017localizing,zhou2018towards,sener2015unsupervised} have also been applied to learn segmentation. Related `set-supervised' learning \cite{richard2018action,li2020set,fayyaz2020set} only assumes the actions in the video are known, but not the ordering.

Several unsupervised methods have also been proposed~\cite{alayrac2016unsupervised,kukleva2019unsupervised,soomro2017unsupervised,sener2018unsupervised}. Alayrac et al. \cite{alayrac2016unsupervised} learn action segmentation without segmentation supervision, using text in addition to video data. \cite{kukleva2019unsupervised} uses $k$-means clustering to do a time-based clustering of features and the Viterbi algorithm segment the videos based on the clusters. \cite{sener2018unsupervised} uses a GMM to learn a transition model between actions. We propose a fully differentiable unsupervised action segmentation, which works from RGB inputs only.

Several datasets for learning from instructional videos 
have been introduced recently:
Breakfast \cite{breakfast}, 50-salads \cite{salads}, the Narrated Instructional Videos (NIV) \cite{alayrac2016unsupervised}, COIN \cite{coin}, HowTo100m \cite{miech2019howto100m}, CrossTask~\cite{zhukov2019crosstask} and PROCEL \cite{procel}. 

\section{Method}


Our goal is to discover atomic actions from a set of instructional videos, while capturing and modeling their temporal structure. Formally, given a set of videos $\mathcal{V} = \{V^1, V^2,...\}$ of a task or set of tasks, the objective is to learn a model that maps a sequence of frames $V^i = [I_t]_{t=1}^T$ from any video to a sequence of atomic action symbols $[a_t \in \mathcal{O}]_{t=1}^T$ where $\mathcal{O}$ is a set of possible action symbols (we drop the index $i$ for simplicity).





\begin{figure} [t]
    \centering
    \includegraphics[width=1.0\linewidth]{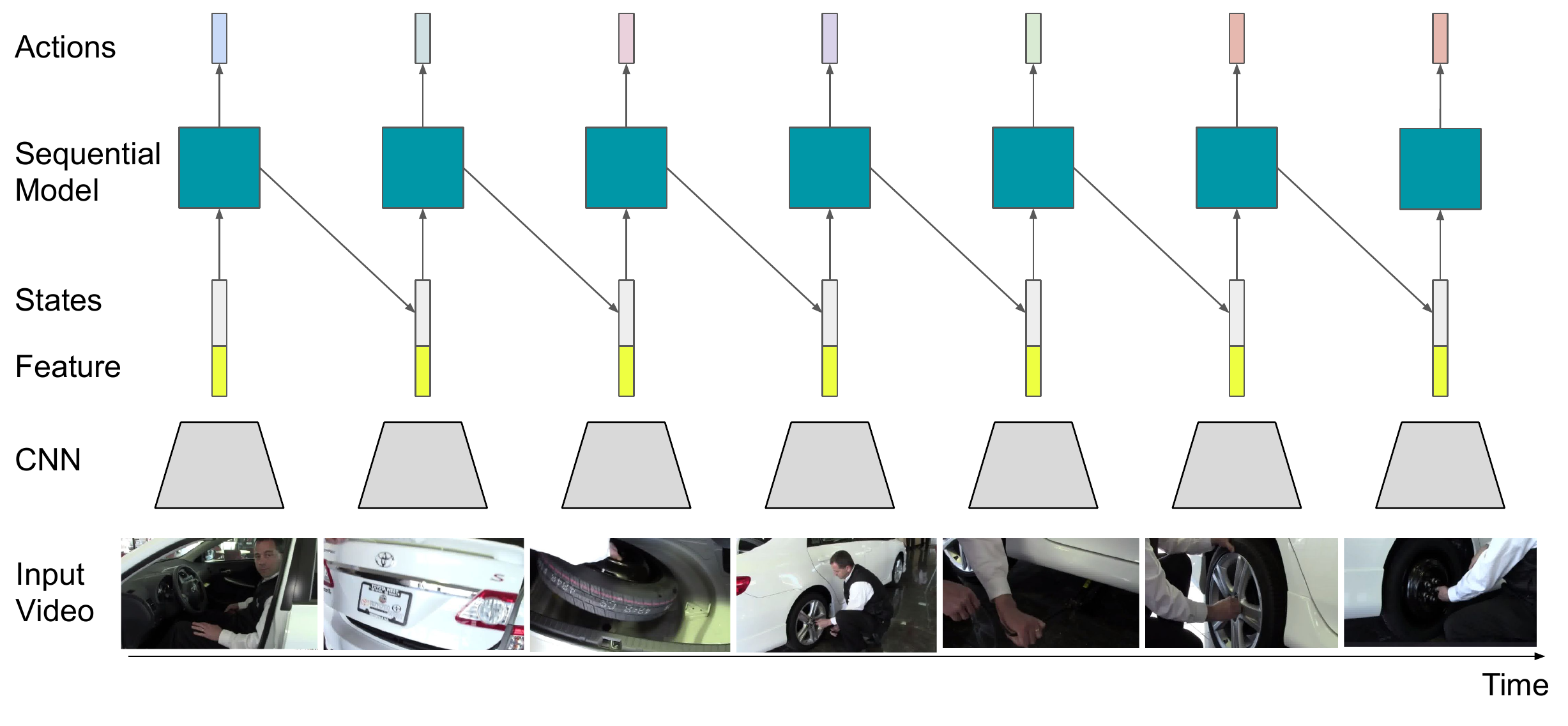}
    \caption{We use a CNN to process each frame, and concatenate those features with the state. Our sequential stochastic model processes each frame, generating a sequence of actions.}
    \label{fig:full-model}
\vspace{-0.3cm}
\end{figure} 

Supervised approaches assume that each frame is labeled with an action, and most weakly supervised approaches assume the actions per video are given in their correct order, but without start and end times. In the unsupervised case, similar to previous works \cite{alayrac2016unsupervised,kukleva2019unsupervised}, we assume no action labels or boundaries are given. To evaluate the approach, we follow the previous setting using the Hungarian algorithm to match predicted actions to ground truth labels. While previous methods used additional data such a subtitles or text \cite{alayrac2016unsupervised}, the proposed approach does not use such information. Our model, however, works with a fixed $k$-the number of actions per task (analogous to setting $k$ in $k$-means clustering), and we run it with a range of values for $k$. This is not a very strict assumption as the number of expected atomic actions per instruction is roughly known,
e.g., about 10 actions for doing CPR, or 40 actions when making a salad. 
For example, a video of making a fried egg will contain the same atomic actions: e.g., cracking the egg, heating a pan, frying the egg, and serving. However, the temporal order, duration and appearance of the actions will vary across videos. 

\subsection{Sequential Stochastic Autoregressive Model}

Our method is based on a sequential stochastic autoregressive model (e.g., \cite{grammar,hinton_sam}). The model consists of three components: $(\mathcal{H}, \mathcal{O}, \mathcal{R})$ where $\mathcal{H}$ is a finite set of states, $\mathcal{O}$ is a finite set of output symbols, and $\mathcal{R}$ is a finite set of transition rules mapping from a state to an output symbol and next state. Importantly, this model is stochastic, i.e., each rule is additionally associated with a probability of being selected, and thus the sum of the rule probabilities for a given state is 1. Note that during training,  $\mathcal{O}$ is just a set of symbols with no semantic meaning or connection to the ground truth labels. For evaluation, following previous works (\cite{kukleva2019unsupervised}), we use the Hungarian algorithm to match these to ground truth symbols.

To implement this method in a differentiable way, we use fully-connected layers and the Gumbel-Softmax trick \cite{jang2016categorical,maddison2016concrete}. Specifically, we use several FC layers taking the current state as input and outputting a vector of logits, representing probabilities of each rule being selected. Next, using the Gumbel-Softmax trick, the model differentiably samples one of the rules. Each time this function is run, a different rule can be selected, learning to generate different sequences. This property is important for the learning of dependencies in sequences. 

Let $G(N_i)$ be this function which maps from the state $N_i \in \mathcal{H}$ to an output symbol (i.e., action) $a\in\mathcal{O}$ and the next state $N_{i+1}$: $(a, N_{i+1}) = G(N_i)$ and $G\in\mathcal{R}$. $N_i$ is a latent vector representation, to be learned through backpropagation. The function $G$ is applied autoregressively to generate a sequence  (Figure~\ref{fig:grammar-overview}). Our approach could be viewed as a state model version of \cite{grammar}, capturing the stochastic sequential structure of the tasks. 


For a full video $V=[I_1, I_2, I_3, \ldots, I_T]$ as input, where each $I_t$ is an RGB image frame from the video, we process the frames by some CNN (e.g., ResNet, I3D \cite{carreira2017quo}, AssembleNet~\cite{Ryoo2020AssembleNet}, we use the latter), resulting in a sequence of feature vectors, $[f_1, f_2, \ldots, f_T]$. These features are used as input to the model, which will generate a sequence of output symbols $S=[a_1,a_2,\ldots, a_T]$ as follows:
\begin{small}
\begin{equation}
\begin{split}
    a_1, N_1 &= G(N_0, f_1),\\
    a_2, N_2 &= G(N_1, f_2),\\
    a_T, N_T &= G(N_{T-1}, f_T)
\end{split}
\end{equation}
\end{small}

The model takes each feature as input and concatenates it with the state which is used as input to $G$ to produce the output. Once applied to every frame, this results in a sequence of actions (Figure~\ref{fig:full-model}). We note that the size of $\mathcal{O}$, $k$, is a hyper-parameter and controls the number of atomic actions expected in the videos. 
We include experiments on the effect of varying the size of $\mathcal{O}$. 

 \begin{figure} 
    \centering
    \includegraphics[width=1.0\linewidth]{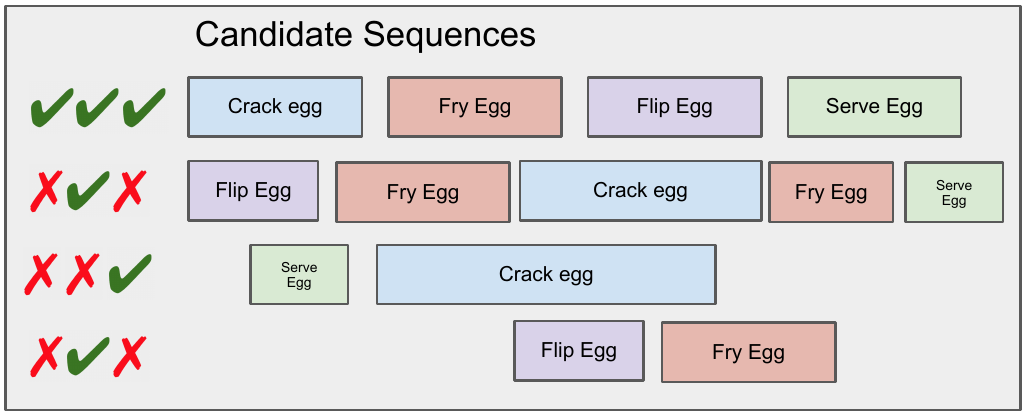}
    \caption{Multiple candidate sequences are generated and ranked. The best sequence according to the ranking function is chosen as the labels for the iteration.}
    \label{fig:candidates}   
\end{figure}


\subsection{Learning by Self-Labeling of Videos}

In order to train the model without ground truth action sequences, we introduce an approach of learning by `self-labeling' videos. The idea is to optimize the model by generating self-supervisory labels that best satisfies the constraints required for atomic actions.
Notably, the stochastic ability to generate multiple sequences is key to this approach.
As a result of the learning, a sequence with better constraint score will become more likely to be generated than the sequences with worse scores.


We first generate multiple candidate sequences, then rank them based on the instructional video constraints, which importantly require no labeled data.
Since the Gumbel-Softmax adds randomness to the model, the output can be different each time $G$ is run with the same input, which is key to the approach. Specifically, the model is run $M$ times, giving $M$ potentially different sequences of actions. We then define a cost function to rank each of the $M$ sequences. The top ranked sequence is selected as the labels which are used for learning. This ranking function constrains the possible generated sequences. 
The ranking function we propose to capture the structure of instructional videos has multiple components:
\begin{itemize}
  \setlength\itemsep{0em}
    \item Every atomic action must occur once in the task. 
    \item Every atomic action should have similar lengths across videos of the same task.
    \item Each symbol should reasonably match the provided visual feature.
\end{itemize}
The best sequence according to the ranking is selected as the action labels for the iteration (Fig.~\ref{fig:candidates}), and the network is trained using a standard cross-entropy loss. We note that depending on the structure of the dataset, these constraints may be adjusted, or others more suitable ones can be designed. In Fig. \ref{fig:example-generations}, we show the top 5 candidate sequences and show how they improve over the learning process.

\textbf{Action Occurrence:}
Given a sequence $S$ of output symbols (i.e., actions), the first constraint ensures that every action appears once. Formally, it is implemented as $C_1(S) = |\mathcal{O}| - \sum_{a\in\mathcal{O}} \mathrm{App}(a)$, where $\mathrm{App}$ is 1 if $a$ appears in $S$ otherwise it is 0.

This computes the number of actions that are \emph{not} predicted as part of the video and is minimized when all actions occur. Similarly, we also penalize sequences that produce the same action multiple disconnected times, as we assume that each video has actions that only appear once (i.e., only break eggs once when frying an egg).  We penalize multiple actions by subtracting the number of disconnected times each
action appears. This constraint is optional, but we include it as it is a property of instructional videos that can be leveraged.

\textbf{Modeling Action Length:}
The constraint ensuring each atomic action has a similar duration across different videos can be implemented in several different ways. The simplest approach is to compute the difference in length compared to the average action length in the video (exact eq. in appendix).

\begin{figure} 
    \centering
    \includegraphics[width=1.0\linewidth]{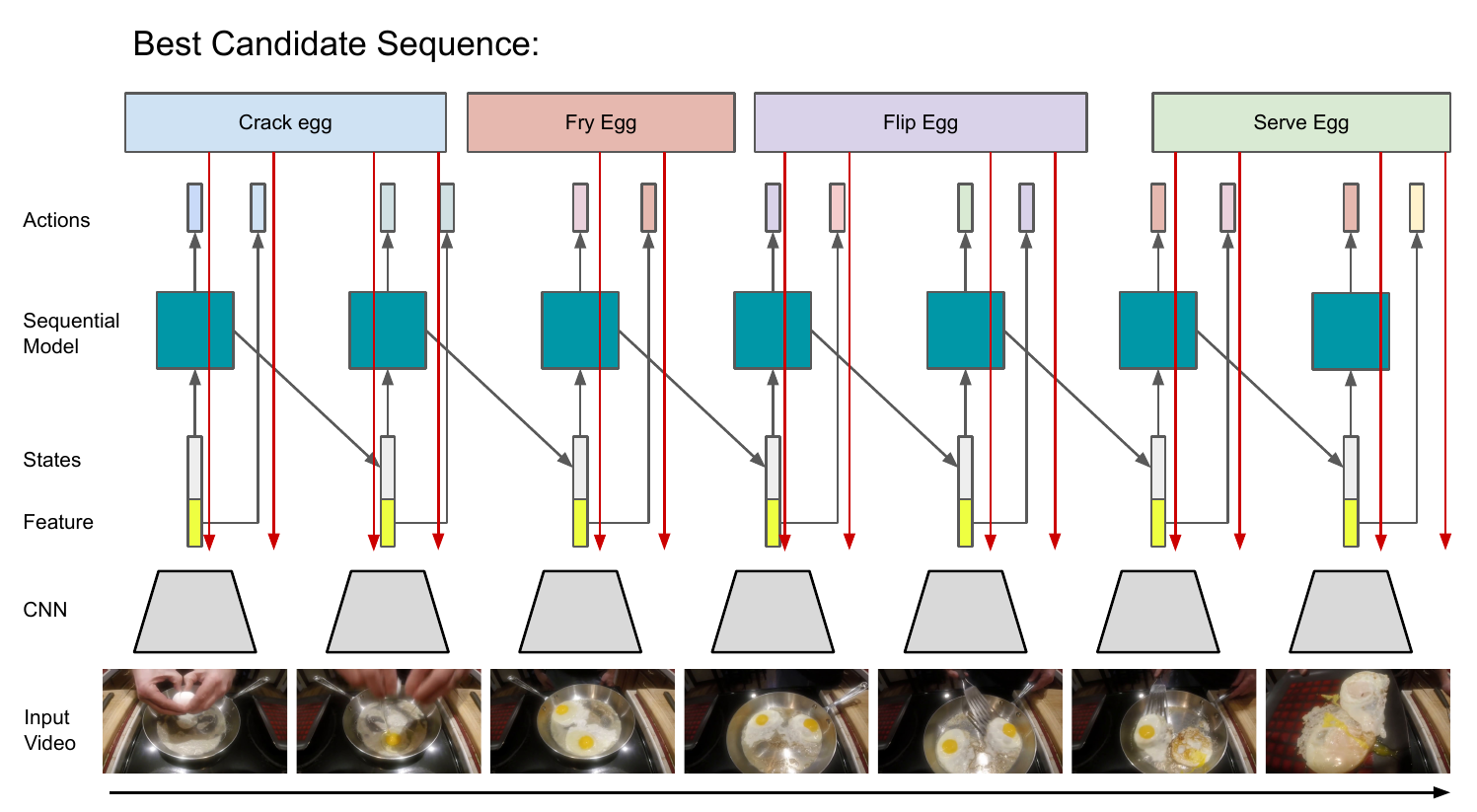}
    \caption{Once the best candidate sequence is selected, it is used to train the model using standard backpropagation. Both the state sequence model as well as the FC-layers generating frame predictions (for $C_3$) are trained.}
    \label{fig:training}
\end{figure}

Another way to model length is by considering the duration of an action to be drawn from a distribution (e.g., Poisson or Gaussian). 
\begin{equation}
\label{eq:cost-len}
    C_2(S) = \sum_{a\in\mathcal{O}} (1-p(L(a, S))),
\end{equation}
where $L(a,S)$ computes the length of action $a$ in sequence $S$ and $p(x) = \frac{\lambda^x\exp{(-\lambda)}}{x!}$ if Poisson or $p(k) = \frac{1}{{\sigma \sqrt {2\pi } }}e^{{{ - \left( {x - \mu } \right)^2 } \mathord{\left/ {\vphantom {{ - \left( {x - \mu } \right)^2 } {2\sigma ^2 }}} \right.} {2\sigma ^2 }}}$ if Gaussian. Since we are minimizing the overall cost function, we use $1-p(x)$ so that it is minimized when the probability is maximal.

The Poisson and Gaussian distributions have parameters: $\lambda$ or $\mu,\sigma$. These parameters control the expected length of the actions in videos. The parameters can be set statically or learned for each action. In the simplest case, we set the parameters to be $\frac{T}{|\mathcal{O}|}$, i.e., the length of each action is the determined by splitting the video equally into actions and $\sigma=1$. In Section \ref{sec:lrn-len}, we detail a method to learn the action length.

\textbf{Modeling Action Probability:}
The third constraint is implemented using the separate classification layer of the network $p(a|f)$, which gives the probability of the frame being classified as action $a$. Formally, $C_3(S) = \sum_{t=1}^T (1-p(a_t|f_t))$, which is the probability that the given frame belongs to the selected action. This constraint is separate from the sequential model and captures independent appearance based probabilities. We note that $a_t$ and $p_t$ are very similar, yet capture different aspects. $p_t$ is generated by a FC-layer applied independently to each frame, while $a_t$ is generated by the
auto-regressive model. We find that using both allows for the creation of the action probability term, which is useful empirically.

We can then compute the rank of any sequence as $C(S) = \gamma_1 C_1(S) + \gamma_2 C_2(S) + \gamma_3 C_3(S)$, where $\gamma_i$ weights the impact of each term. In practice setting $\gamma_2$ and $\gamma_3$ to $\frac{1}{|S|}$ and $\gamma_1=\frac{1}{|\mathcal{O}|}$ works well.

\textbf{Learning Actions:}
To choose the self-labeling, we sample $K$ sequences, compute each cost and select the sequence that minimizes the above cost function. This gives the best segmentation of actions (at this iteration of labeling) based on the defined constraints. 
\begin{equation}
    \hat{S} = \texttt{argmin}_S C(S).
\end{equation}

We note that this cost function does not need to be differentiable. The cost function is only used to choose the self-labels. Once the labels are selected, the standard cross-entropy loss function with backpropagation is used to train the model. The cost function gives a strong prior for how to choose the labels without any annotations, and allows unsupervised learning. Formally, given the selected labels $\hat{S} = [\hat{a}_1, \hat{a}_2, \ldots \hat{a}_T]$  i.e., they can now serve as a weak ground truth at this iteration, the output of the model $A = [a_1, a_2, \dots a_T]$, and the outputs of the classification layer $P = [p_1, p_2, \ldots p_T]$, where $a_t$ and $p_t$ are probability vectors for each action, we define the loss as:
\begin{equation}
\label{eq:loss}
    \mathcal{L}(\hat{S}, A, P) =  - \sum_{i\in\mathcal{O}} \sum_{t=1}^T \hat{a}_{t,i} \log(a_{t,i}) + \hat{a}_{t,i}\log(p_{t,i}).
\end{equation}
This loss trains both the classification layer as well as the model.

We also allow a null class to indicate that no actions are occurring in the given frames. This class is not used in any of the above constraints, i.e., it can occur wherever it wants, for as long as needed and as many times as needed. We omit frames labeled with the null class when calculating the cost function and find that the constraint encouraging each action to occur once eliminates the solution where only the null class is chosen.

\subsection{Cross-Video Matching}
The above constraints work reasonably well for a single video, however when we have multiple videos with the same actions, we can further improve the ranking function by adding a cross-video matching constraint. The motivation for this is that while breaking an egg can be visually different between two videos (different bowls, camera angles and background), the action, e.g., overall motion and object, are the same. 

To encourage the model to learn and use this, especially in videos having partially ordered sequences (e.g., break egg, heat pan vs. heat pan, break egg), we add a cross-video matching term. Given a video segment the model labeled as an action $f_a$ from one video, a segment $\hat{f}_a$ the model labeled as the same action from a second video, and a segment $f_b$ the modeled labeled as a different action from any video, we can measure the cross-video similarity using standard methods, such as a triplet loss
\begin{equation}
    \label{eq:triplet}
    \mathcal{L}_T (f_a, \hat{f}_a, f_b) = ||f_a - \hat{f}_a||_2 - ||f_a - f_b||_2 + \alpha,
\end{equation}
or a contrastive loss
\begin{equation}
    \label{eq:contrastive}
    \mathcal{L}_C (f_a, \hat{f}_a, f_b) = \frac{1}{2}||f_a - \hat{f}_a||_2 + \frac{1}{2}\max{(0, \alpha - ||f_a - f_b||_2)}.
\end{equation}

These two functions capture similar properties but in slightly different ways. The triplet loss maximizes the difference between anchor (e.g., $f_a$) and positive ($\hat{f}_a$)/negative ($f_b$) distance. While the contrastive loss maximizes the distance between $f_a$ and $f_b$ separately from minimizing the distance between $f_a$ and $\hat{f}_a$. This results in slightly different cross-video matching metrics.

As these functions are differentiable, we can directly add this to the loss function (Eq. \ref{eq:loss}) or to the cost function (Eq. \ref{eq:cost-len}) or both. By adding this to the cost function, we are ensuring that the chosen labeling of the videos is most consistent for feature representations. By adding it to the loss function, we are encouraging the learned representations to be similar for the actions with the same selected labels and different for other actions. We analyze the effect of these in Table \ref{tab:cross-video}.

\begin{figure}
    \centering
    \includegraphics[width=\linewidth]{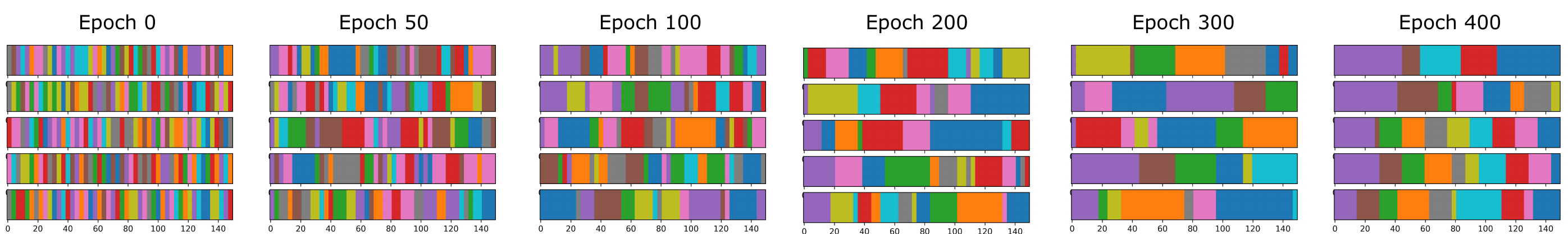}
    \caption{Candidate sequences at different stages of training. The sequences shown are the top 5 ranked sequences (rows) at the given epoch. 
    The top one is selected as supervision for the given step. The colors represent the discovered action (with no labels).}
    \label{fig:example-generations}
\end{figure}

\subsection{Self-labeling Training Method}
Using the previous components, we now describe the full training method, which follows an EM-like procedure. In the first step, we find the optimal set of action self-labels given the current model parameters and the ranking function. In the second step, we optimize the model parameters (and optionally some ranking function parameters) for the selected self-labeling (Figure~\ref{fig:training}). After taking both steps, we have completed one iteration. 
Following standard neural network training, we do each step for a mini-batch of 32 samples. The model is trained for 500 epochs. Due to the iterative update of the labels at each step, we observe that this method requires more epochs than supervised learning. We use gradient descent with momentum to optimize the network parameters with a learning rate set to 0.1 following a cosine decay schedule.


\begin{table}
\centering
    \begin{tabular}{l|c}
    \toprule
    Method & F1 score  \\
    \midrule
    \multicolumn{2}{l}{\textbf{Supervised Baselines}}\\
    \midrule
     VGG~\cite{simonyan2014very}, from Alayrac et al.~\cite{alayrac2016unsupervised} &  0.376 \\
    I3D, Carreira et al. \cite{carreira2017quo} & 0.472  \\
    AssembleNet, Ryoo et al. \cite{Ryoo2020AssembleNet} & 0.558 \\
    \midrule
    \multicolumn{2}{l}{\textbf{Weakly-supervised}}\\
    \midrule
    CTC, Huang et al. \cite{huang2016connectionist} +AssembleNet~\cite{Ryoo2020AssembleNet} & 0.312  \\
    ECTC, Huang et al. \cite{huang2016connectionist} +AssembleNet~\cite{Ryoo2020AssembleNet}  & 0.334 \\
    \midrule
    \multicolumn{2}{l}{\textbf{Unsupervised}}\\
    \midrule
    Uniform Sampling & 0.187 \\
    Alayrac et al. \cite{alayrac2016unsupervised} & 0.238 \\
    Kukleva et al \cite{kukleva2019unsupervised} & 0.283 \\
    JointSeqFL, Elhamifar et al. \cite{procel} & 0.373 \\
    Ours & \textbf{0.457}   \\
    \bottomrule
    \end{tabular}
 \centering
    \caption{Results on the NIV dataset}
    \label{tab:cvpr16-main-results}
\vspace{-0.1cm}  
\end{table}

\textbf{Learning action length:}
\label{sec:lrn-len}
As an optional training phase, we update some parameters of the ranking function. The particular parameters to learn are those determining the length of each action, since some atomic actions will be longer than others and we often do not know the actual length of the action. To do this, we modify the length model so that it has a $\lambda_a$ or $\mu_a, \sigma_a$ to represent the length of each action $a$. To estimate these values, after the backpropagation of the gradients, we run the model in inference mode to obtain a segmentation of the video. For each action, we then compute its average length (and optionally variance) which we can use to update $\lambda_a$ or $\mu_a, \sigma_a$.


\textbf{Segmenting a video at inference:} CNN features are computed for each frame and the learned model is applied on those features. During rule selection, we greedily select the most probable rule. Future work can improve this by considering multiple possible sequences (e.g., following the Viterbi algorithm).

\begin{table}
    %
    %
        \centering
    \begin{tabular}{l|ccc}
    \toprule
    Method & MoF\\
    \midrule
    \multicolumn{2}{l}{\textbf{Supervised Baselines}}\\
    \midrule
         VGG~\cite{simonyan2014very}, from Alayrac et al.~\cite{alayrac2016unsupervised} &   60.8 \\
        I3D, Carreira et al. \cite{carreira2017quo} & 72.8\\
    AssembleNet, Ryoo et al.~\cite{Ryoo2020AssembleNet} & 77.6  \\
    \midrule
    \multicolumn{2}{l}{\textbf{Weakly-supervised}}\\
    \midrule
    CTC, Huang et al. \cite{huang2016connectionist} & 11.9 \\
    HTK, Kuehne et al. \cite{kuehne2017weakly} & 24.7 \\
    HMM + RNN, Richard et al. \cite{richard2017weakly} & 45.5 \\
    NN-Viterbi, Richard et al. \cite{richard2018neuralnetwork} & 49.4 \\
    Ours, weakly supervised \footnotemark  & 53.7 \\
    \midrule
    \multicolumn{2}{l}{\textbf{Unsupervised}}\\
    \midrule
    Kukleva et al \cite{kukleva2019unsupervised} & 30.2 \\
    Ours & \textbf{39.7} \\
    \bottomrule
    \end{tabular}
    \caption{Results on the 50-salads dataset.}
    \label{tab:salads-main-results}
 \vspace{-0.2cm}
    \end{table}

\section{Experiments}
\footnotetext{For the weakly-supervised setting, we use activity order as supervision, equivalent to previous works.}

We evaluate our unsupervised atomic action discovery approach on multiple video segmentation datasets, confirming that our self-generated action annotations form meaningful action segments.
We note that there is only a handful of methods that have attempted unsupervised activity segmentation. 
Thus, we also compare to several fully-supervised methods and to weakly-supervised ones. 

\noindent\textbf{Datasets:} We compare results on the \textbf{50-salads dataset} \cite{salads}, which contains 50 videos of people making salads (i.e., a single task). The videos contain the same set of actions (e.g., cut lettuce, cut tomato, etc), but the ordering of actions is different in each video. We compare on \textbf{Narrated Instructional Videos (NIV) dataset} \cite{alayrac2016unsupervised}, which contains 5 different tasks (CPR, changing a tire, making coffee, jumping a car, re-potting a plant). These videos have a more structured order. Finally, we use the \textbf{Breakfast dataset} \cite{breakfast} which contains videos of people making breakfast dishes from various camera angles and environments. 
We chose these datasets as they cover a wide variety of approaches focused on unsupervised, weakly-supervised, and fully-supervised action segmentation, allowing a comparison to them. 
Furthermore, these datasets, unlike other related ones, provide appropriate annotations for the evaluation of atomic action learning.

\noindent\textbf{Evaluation Metrics:} 
We follow all previously established protocols for evaluation in each dataset. We first use the Hungarian algorithm to map the predicted action symbols to action classes in the ground truth. Since different metrics are used for different datasets we report the previously adopted metrics per dataset. 
Specifically, for NIV, we predict a temporal interval for each action, then compute the F1 score if the interval falls within a ground truth interval (following \cite{alayrac2016unsupervised}). For 50-salads, we compute the mean-over-frames (MoF) which is the per-frame accuracy for each frame. For Breakfast, we report both the MoF and Jaccard measure, following previous works \cite{huang2016connectionist,richard2017weakly,kukleva2019unsupervised}.

\subsection{Comparison to the state-of-the-art}
In Tables \ref{tab:cvpr16-main-results}, \ref{tab:salads-main-results}, \ref{tab:brkfst-main-results}, we compare our approach to previous state-of-the-art methods.
While there are few works on the fully unsupervised case, we note that our approach, together with strong video features, provides better segmentation results than previous unsupervised and even weakly-supervised methods (JointSeqFL \cite{procel}  uses optical flow and does not provide results on 50-salads or Breakfast).


For full comparison we include strong supervised baselines, e.g., I3D \cite{carreira2017quo}, and AssembleNet~\cite{Ryoo2020AssembleNet}. We also use implementations of the CTC \cite{ctc} and ECTC \cite{huang2016connectionist} methods using the AssembleNet backbone~\cite{Ryoo2020AssembleNet}. 
Our unsupervised approach outperforms many weakly-supervised ones too (Tables~\ref{tab:cvpr16-main-results},~\ref{tab:brkfst-main-results}).

\noindent\textbf{Qualitative Analysis} In Fig. \ref{fig:example-generations}, we show the generated candidate sequences at different stages of learning. It can be seen that initially the generated sequences are entirely random and over-segmented. As training progresses, the generated sequences start to match the constraints. After 400 epochs, the generated sequences show similar order and length constraints, and better match the ground truth (as shown in the evaluation).
Figures~\ref{fig:examples},~\ref{fig:varying-number} show example results of our method.

\begin{table}    
    %
    %
        \centering
 \small
    \begin{tabular}{l|cc}
    \toprule
    Method &  MoF & Jaccard \\
    \midrule
    \multicolumn{2}{l}{\textbf{Supervised Baselines}}\\
    \midrule
        VGG~\cite{simonyan2014very}, from Alayrac et al.~\cite{alayrac2016unsupervised} &  62.8 & 75.4 \\
        I3D, Carreira et al.\cite{carreira2017quo} & 67.8 & 79.4 \\
        AssembleNet, Ryoo et al.~\cite{Ryoo2020AssembleNet} & 72.5 & 82.1 \\
    \midrule
    \multicolumn{2}{l}{\textbf{Weakly-supervised}}\\
    \midrule
    OCDC, Huang et al. \cite{huang2016connectionist} & 8.9 & 23.4 \\
    ECTC, Huang et al.\cite{huang2016connectionist} & 27.7 & - \\
    HMM + RNN, Richard et al.
    \cite{richard2017weakly} & 33.3 & 47.3 \\
    \midrule
    \multicolumn{2}{l}{\textbf{Unsupervised}}\\
    \midrule
    SCV, Li and Todorovic
    \cite{li2020set} & 30.2 & - \\
    SCT, Fayyaz and Gall
    \cite{fayyaz2020set} & 30.4 & - \\
    Sener et al \cite{sener2018unsupervised} & 34.6 & 47.1 \\
    Kukleva et al \cite{kukleva2019unsupervised} & 41.8 & - \\
    Ours & \textbf{43.5} & \textbf{54.4} \\
    \bottomrule
    \end{tabular}
           \caption{Results on the Breakfast dataset.}
    \label{tab:brkfst-main-results}
    \vspace{-0.5cm}
\end{table}

\subsection{Ablation experiments}


\noindent\textbf{Effect of sequential models for weak-supervision.}
We conduct a set of experiments to determine the effect of learning temporal information in different ways. The CTC loss, which bases the loss on the probability of the sequence occurring \cite{ctc}, can be applied directly on per-frame features, without any underlying RNN or temporal model. In Table \ref{tab:temporal-exp}, we compare the effect of using the CTC loss with per-frame CNN features, an RNN, and our model. We note that our model is an RNN with a restricted, discrete set of states, but is able to stochastically select states. We find that adding temporal modeling to the CNN features is beneficial, and for these datasets, our model further improves performance. These experiments all use order-of-activity labels, and are weakly-supervised. They also all use AssembleNet.

\begin{table}[]
    \centering
     \begin{tabular}{l|ccc}
    \toprule
    Method & NIV & 50-Salads & Breakfast \\
    \midrule
    Supervised     & 0.558 & 77.6 & 72.5 \\
    \midrule
    CTC     & 0.312 & 42.8 & 38.7  \\
    RNN + CTC & 0.388 & 47.9 & 42.4 \\
    Ours + CTC & \textbf{0.480} & \textbf{52.8} & \textbf{45.3} \\
    \bottomrule
    \end{tabular}
       \caption{Comparing different weakly-supervised models. All using AssembleNet features. The supervised counterpart at the top.}
    \label{tab:temporal-exp}
\end{table}

\begin{table}[]
    %
    %
       \centering
       \small
    \begin{tabular}{l|cc}
    \toprule
    Cost & 50-Salads & Brkfst \\
    \midrule
    Randomly pick candidate & 12.5 & 10.8 \\
    No Gumbel-Softmax & 10.5 & 9.7 \\
    \midrule
    Occurrence ($C_1$)   & 22.4  & 19.8 \\
    Length ($C_2$)    & 19.6 & 17.8 \\
    $p(a|f)$ ($C_3$) & 21.5 & 18.8 \\
    \midrule
    $C_1 + C_2$ & 27.5 & 25.4 \\
    $C_1 + C_3$ & 30.3 & 28.4 \\
    $C_2 + C_3$ & 29.7 & 27.8 \\
    $C_1 + C_2 + C_3$ & 33.4 & 29.8 \\
    \bottomrule
    \end{tabular}
   \caption{Ablation with cost function terms$^2$}
    \label{tab:cost_fn}
 
\end{table}

\begin{table}[]
    \centering
    \small
  
    \begin{tabular}{lc|ccc}
    \toprule
    Function & Use & NIV & 50-Salads & Brkfst \\
    \midrule
    None & N/A & 0.420 & 33.4 & 31.7 \\
    triplet & cost & 0.485 & 37.8 & 37.8 \\
    contr. & cost & 0.478 & 37.9 & 36.4\\
    triplet & loss & 0.492 & 38.4 & 37.5 \\
    contr. & loss & 0.478 & 39.2 & 38.4 \\
    triplet & both & 0.442 & 35.7 & 36.9 \\
    contr. & both & 0.448 & 36.2 & 35.2 \\
    \bottomrule
    \end{tabular}
   \caption{Cross video matching$^3$}    
   \label{tab:cross-video} 
   \end{table}
   
\begin{table} []
    %
    %
    \centering
    \small
    \begin{tabular}{lc|ccc}
    \toprule
    Method & Learned & NIV & 50-Salads & Brkfst \\
    \midrule
        Avg. & no & 0.420 & 33.4 & 31.7  \\
        Gaussian & no & 0.418 & 32.8 & 34.5 \\
        Poisson & no & 0.435 & 33.6 & 32.8 \\
        Gaussian & yes & 0.432 & 35.7 & 36.5 \\
        Poisson & yes & 0.447 & 37.9 & 37.9 \\
    \bottomrule
    \end{tabular}
   \caption{Different length models, including learned and fixed lengths per action.$^3$}
    \label{tab:length-model}
    \vspace{-6mm}
\end{table}

\begin{table}[]
    \centering
    \footnotesize
    \begin{tabular}{l|ccccc|c}
    \toprule
 Method & chng  & CPR & repot  & make  & jump & Avg. \\
        &  tire &  & plant & coffee & car & \\
 \midrule
Alaryac et al. \cite{alayrac2016unsupervised}  & 0.41 & 0.32 & 0.18 & 0.20 & 0.08 & 0.238 \\
Kukleva et al. \cite{kukleva2019unsupervised} &-&-&-&-&-& 0.283\\
Ours VGG & 0.53 & 0.46 & 0.29 & 0.35 & 0.25 & 0.376 \\
Ours AssembleNet & 0.63 & 0.54 & 0.381 & 0.42 & 0.315 & 0.457 \\
\bottomrule
    \end{tabular}
    \caption{Comparison on the NIV dataset of the proposed approach on VGG and AssembleNet features.}
    \label{tab:features}
    \vspace{-4mm}
\end{table}

\begin{figure*}
    \centering
    \includegraphics[width=0.19\linewidth,height=2cm]{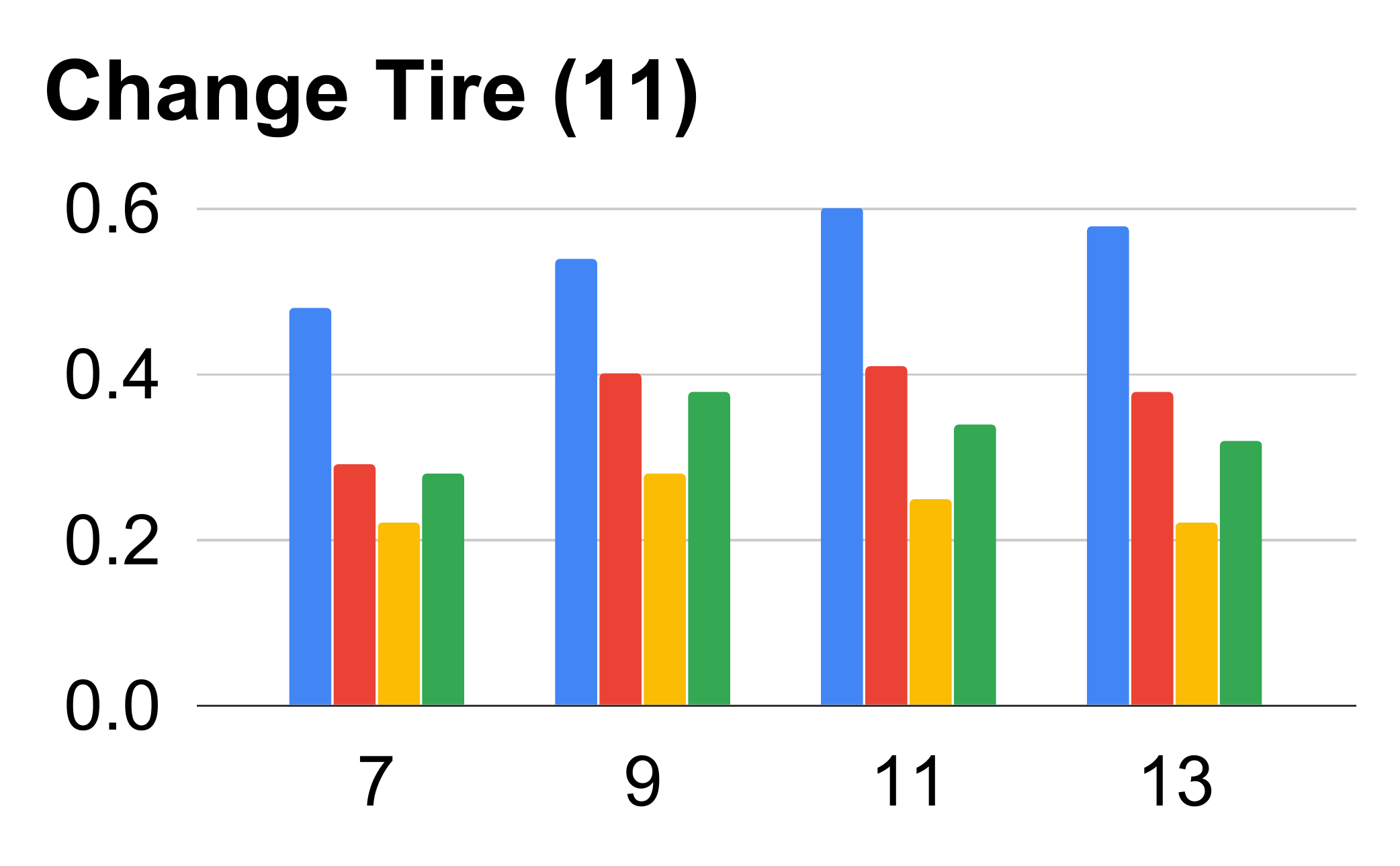}
    \includegraphics[width=0.19\linewidth, height=2cm]{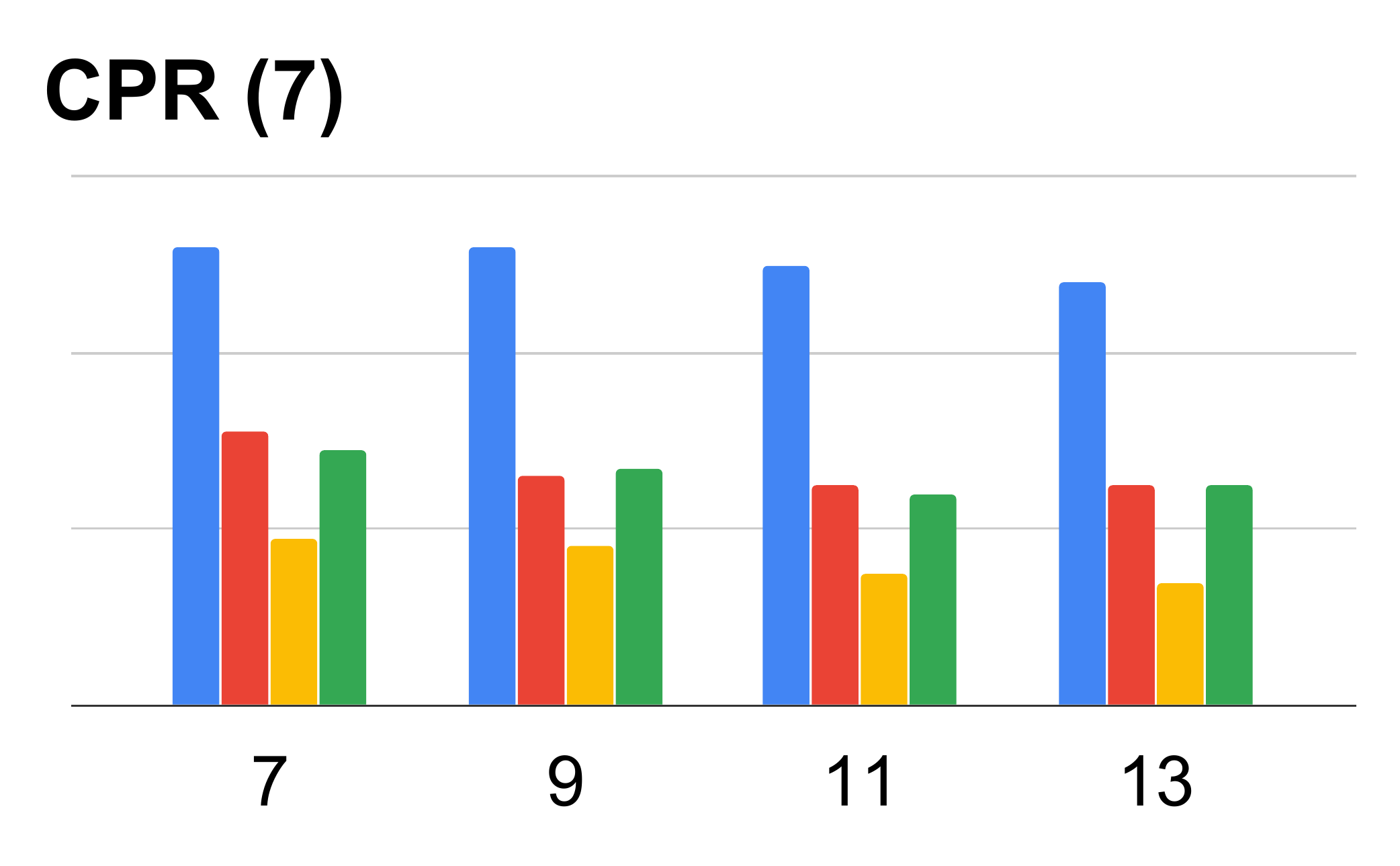}
    \includegraphics[width=0.19\linewidth, height=2cm]{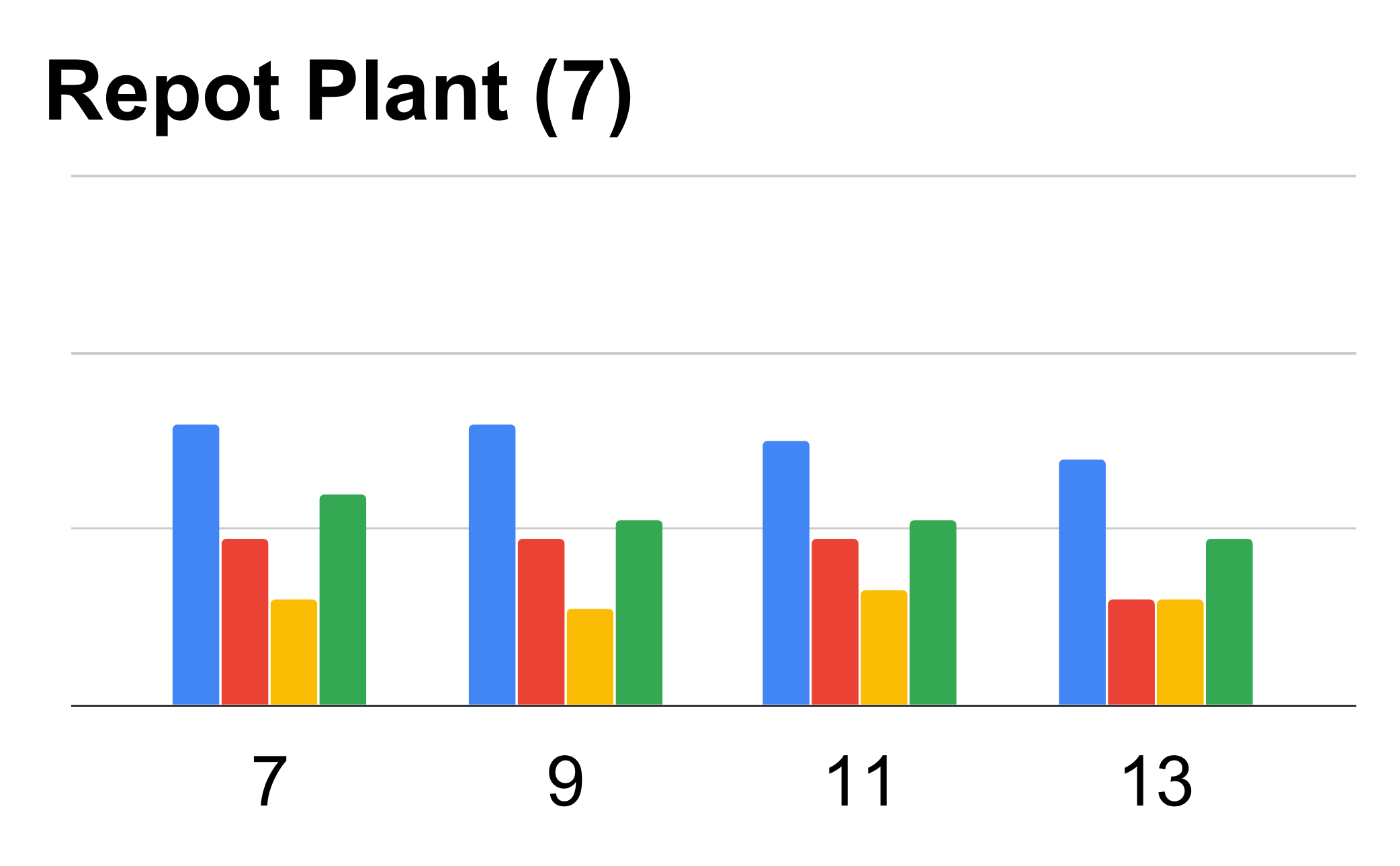}
    \includegraphics[width=0.19\linewidth, height=2cm]{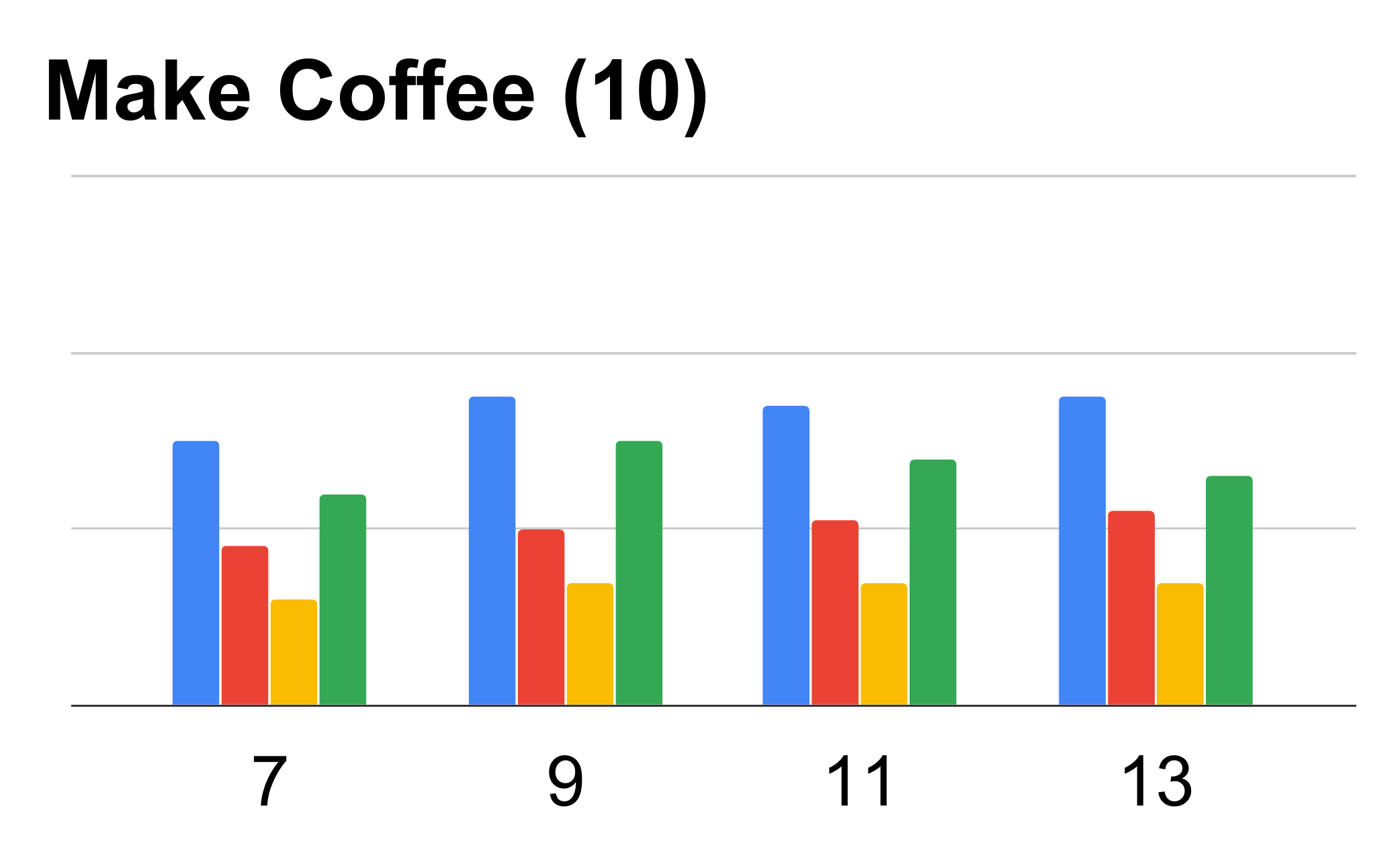}
    \includegraphics[width=0.19\linewidth, height=2cm]{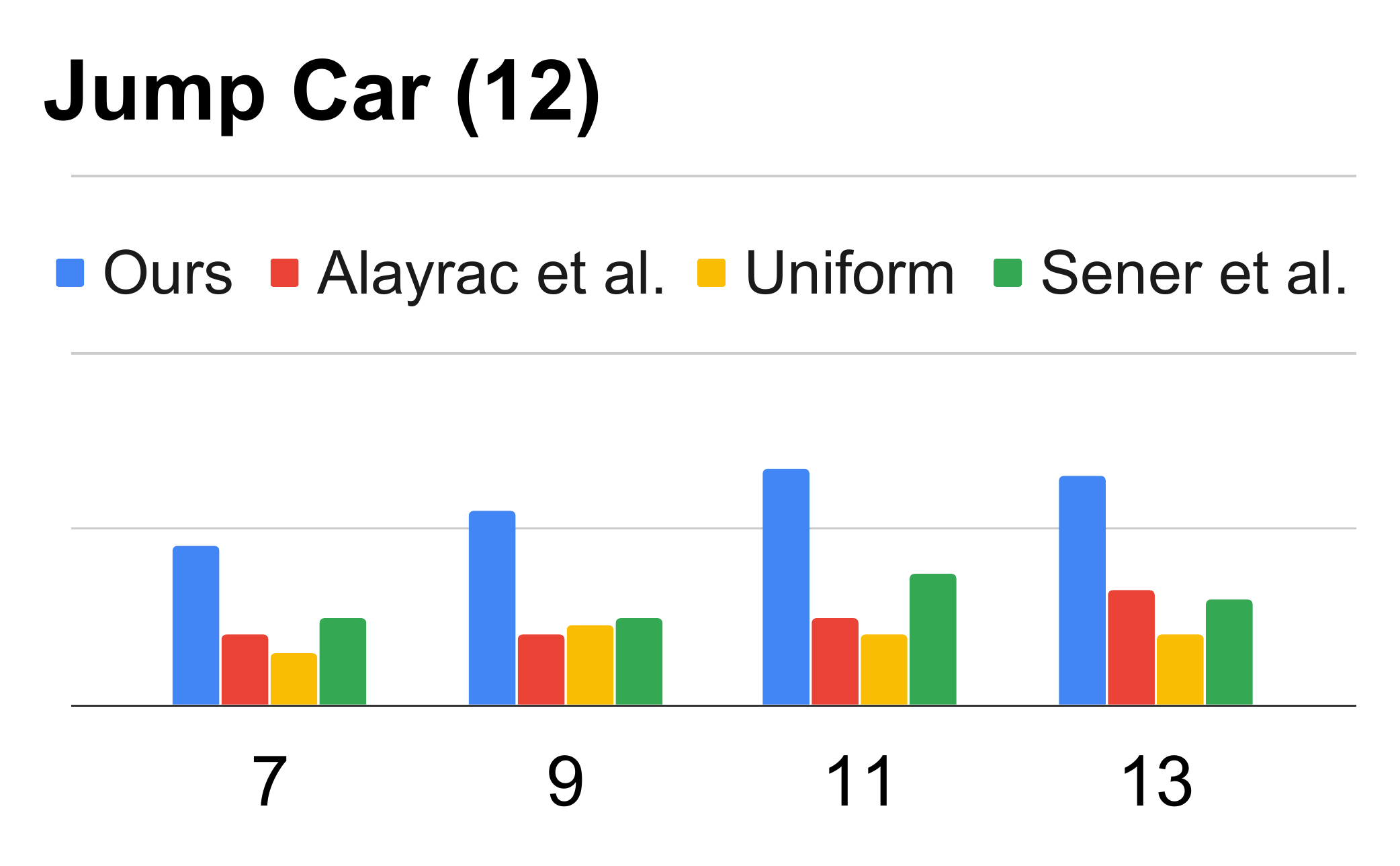}
    \caption{F1 value for varying the number of actions used in the model, compared to prior work. The number in parenthesis indicates the ground-truth number of actions for each activity. Full results are in the sup. materials.}
    \label{fig:num-steps}
\end{figure*}

\noindent\textbf{Effects of the cost function constraints.}
To determine how each cost function impacts the resulting performance, we compare various combinations of the terms. The results are shown in Table \ref{tab:cost_fn}. We find that each term is important to the self-labeling of the videos\footnote{These ablation methods do not use our full cross-video matching or action duration learning, thus the performances are slightly lower than the our best results.}. Generating better self-labels improves model performance, and each component is beneficial to the selection process. Intuitively, this makes sense, as the terms were picked based on prior knowledge about instructional videos. We also compare to random selection of the candidate labeling and a version without using the Gumbel-Softmax. Both alternatives perform poorly, confirming the benefit of the proposed approach.

\addtocounter{footnote}{1}
\footnotetext{To isolate the effect, Table \ref{tab:cross-video} uses the length model without learning, and Table \ref{tab:length-model} uses no cross video matching.}

\noindent\textbf{Methods for cross-video matching.}
In Table \ref{tab:cross-video}, we compare the results for the different methods of cross-video matching on the 50-salads dataset. We compare both the triplet loss (Eq. \ref{eq:triplet}) and the constrastive loss (Eq. \ref{eq:contrastive}) using them as part of the cost function, training loss function or both. We find that using the contrastive as part of the training loss performs the best, as this further encourages the learned representation to match the chosen labels.

\noindent\textbf{Methods for length modeling.}
In Table \ref{tab:length-model}, we compare the different methods to model the length of each action. We find that learning the length of each action (Section \ref{sec:lrn-len}) is most beneficial.

\begin{figure}
    \centering
    \includegraphics[width=\linewidth]{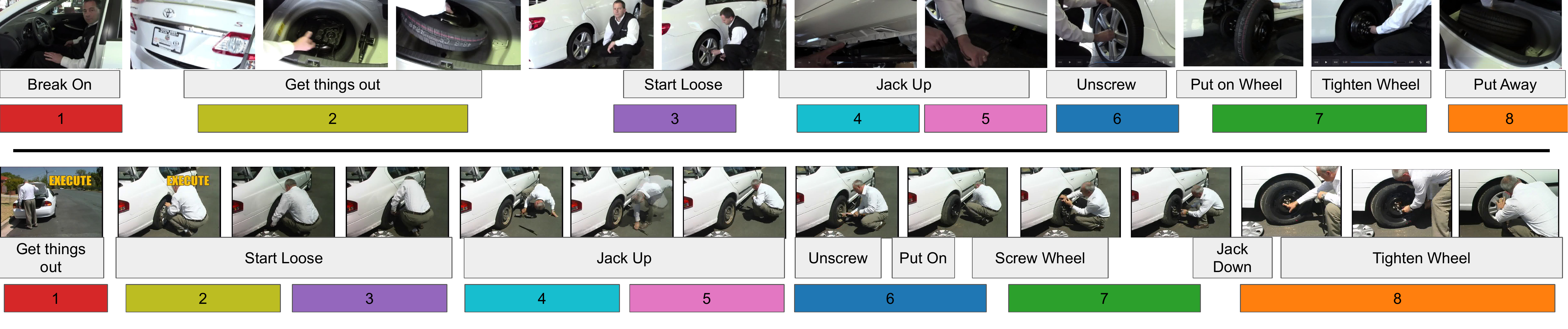}
    \caption{Two example videos from the `change tire' activity. The ground truth is shown in grey, the model's top rank segmentation is shown in colors. NIV dataset.}
    \label{fig:examples}
\end{figure}

\begin{figure}
    \centering
    \includegraphics[width=\linewidth]{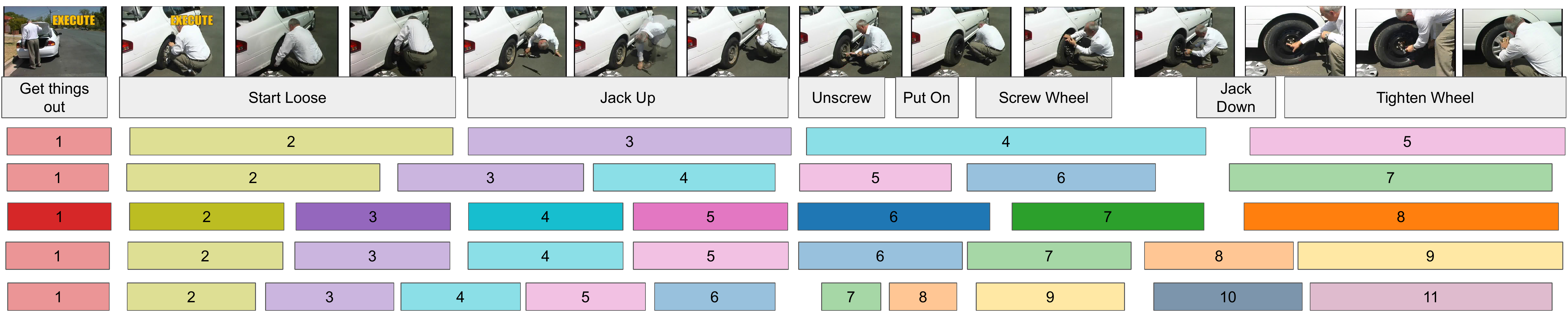}
    \caption{Example segmentation of the `change tire' activity varying the number of actions from 5 to 11. The segmentations generally match, even when the number of actions does not match the ground truth number.}
    \label{fig:varying-number}
\end{figure}

\noindent\textbf{Varying the number of actions.}
As $\mathcal{O}$ is a hyper-parameter controlling the number of actions to segment the video into, we conduct experiments on NIV varying the number of actions/size of $\mathcal{O}$ to evaluate the effect this hyper-parameter has. The results are shown in Figure \ref{fig:num-steps}. Overall, we find that the model is not overly-sensitive to this hyper-parameter, but it does have some impact on the performance due to the fact that each action must appear at least once in the video.

\noindent\textbf{Feature comparisons.}
As our work uses AssembleNet \cite{Ryoo2020AssembleNet} features, in Table \ref{tab:features} we compare the proposed approach to previous ones using both VGG and AssembleNet features. As shown, even using VGG features, our approach outperforms previous methods.

\subsection{Using oracles}
To better understand where the model succeeds and fails, we compare effects of adding different oracle information.

\begin{figure}
\centering
    \includegraphics[width=0.7\linewidth]{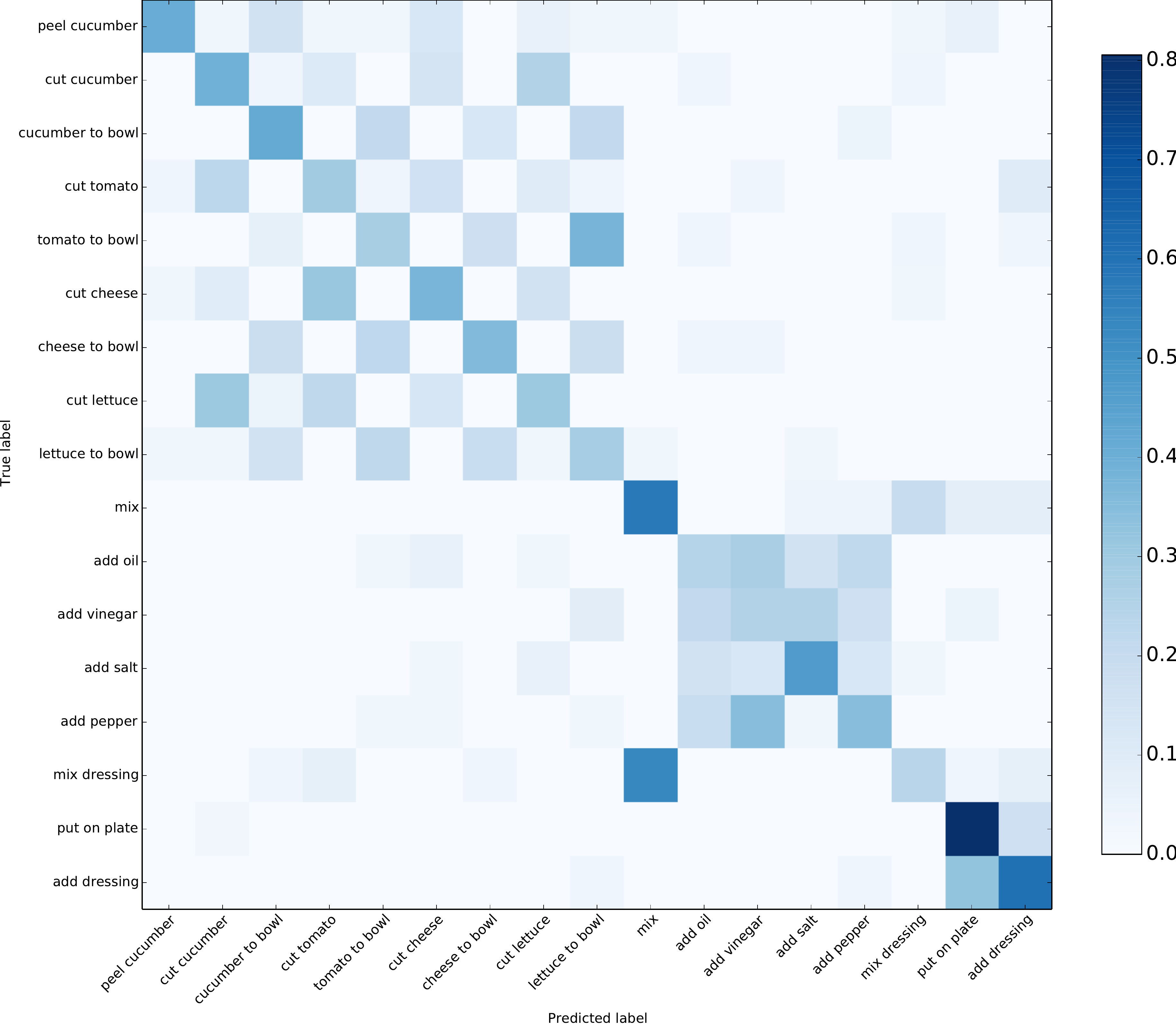}
    \caption{Confusion matrix for 50 salads dataset. Most of the confusion is around the objects, e.g., `cut lettuce' vs. `cut tomato.'}
    \label{fig:confusion}
\vspace{-0.4cm}
\end{figure}

We compare our model using 5 different oracles. (1) Object oracle that tells us the object of interest for each frame. For example, if cutting cucumber is the action, cucumber is the given object; for frying an egg, egg is given. (2) Action oracle (e.g., cut, peel, mix, etc.). (3) High-level action oracle, i.e., a grouping of several relation actions. For example: `prepare salad' which contains cut, peel, add to bowl actions. (4) Time oracle which gives the start and end times of each action, but not the action itself. (5) Order oracle: gives the previous and next action, but not the current action (only usable for classifying the current frame). 

The results are shown in Table \ref{tab:oracles} for the 50-salads dataset.
We find that the model performs quite well in learning the temporal ordering and structure of the data, as the performance only slightly increases when including additional temporal information. Adding perfect object information greatly boosts performance, suggesting that the current model struggles to learn objects.

\begin{table}
    \centering
        \begin{tabular}{l|c}
    \toprule
   Oracle & Accuracy \% \\
   None &33.4\\
   Action & 39.8\\
   High-level action & 34.8\\
   Temporal &  36.8\\
   Ordering & 39.4 \\
     Object & 48.5 \\
    \bottomrule
    \end{tabular}
   \centering
    \caption{Comparison of different oracles on 50-salads}
    \label{tab:oracles}
\end{table}

\begin{table}
   \centering
  
\begin{tabular}{l|ccccccc}
    \toprule
    
\# labeled &  0 & 1 &2 & 3 & 4 & 5 & 50\\
Acc. \% &  33.4 & 42.8 &44.3 & 45.2 & 46.6 & 47.1 & 77.6\\

    \bottomrule
    \end{tabular}
      \caption{Classification accuracy for different number of labeled examples (50 labels means all examples are labeled). 50-salads.}
    \label{tab:num-labeled}
\end{table}

    
    
    
    %
    %

Figure~\ref{fig:confusion} shows the confusion matrix for the 50 Salads dataset. As seen, actions are well separated from one another. There is confusion among objects (top left portion), e.g., `cut cucumber', `cut tomato' and `cut lettuce' are confused, but actions, e.g., `cut' and `peel' are well separated. This confirms actions are well understood by the model.

\noindent\textbf{Weak Labeling Oracle.}
Here we have an oracle that gives $N$ true examples and the model `mines' the action from the other videos. This allows further analysis of the impact of unsupervised learning. We conduct a set of experiments comparing the unsupervised approach against $N$ fully-labeled videos given. $N$ videos are selected at random for supervised learning. The we perform the iterative, unsupervised training method for the remaining videos. The results are averaged over 10 different runs, each with a different set of labeled videos. Table \ref{tab:num-labeled} shows the results. We find that adding one true video greatly boosts performance (+9\%), and each additional video adds only about 1\% to fully supervised performance, showing the strong benefit of the self-labeling approach.

\vspace{-6pt}
\section{Conclusions and future work}
\vspace{-6pt}
We present a novel approach for unsupervised action segmentation 
for instructional videos. Based on a stochastic autoregressive model and ranking function, the algorithm is able to learn to  self-label and segment actions without supervision. 
Our approach outperforms the unsupervised methods, in some cases weakly supervised too.

{\small
\bibliographystyle{ieee_fullname}
\bibliography{bib}
}

\clearpage
\newpage
\appendix

\section{Implementation Details}
The model is implemented in PyTorch. The pretrained models are on Kinetics-600 with overlapping classes removed, as is standard practice for unsupervised approaches (see Section 3 below for all removed classes). As base networks, which are needed to obtain initial features from the videos, we use and compare VGG \cite{simonyan2014very}, I3D \cite{carreira2017quo} and AssembleNet \cite{Ryoo2020AssembleNet}. These cover a wide range of networks previously used for video understanding (e.g. I3D), for unsupervised video segmentation, where VGG is often used, and current state-of-the-art models (AssembleNet).
Our main model uses the AssembleNet backbone, which contains ResNet blocks of interleaved spatial and 1d temporal convolutions. It is equivalent in the number of parameters to a ResNet-50 (2+1)D network.

We used all three constraints (C1, C2, C3) with the weights set as described in Section 3.2 of the main paper. We used cross-video matching in the loss function with the triplet loss formulation. We used the learned Poisson version of length modeling. These corresponded to the best values in each of Tables 5-7 of the paper.

During evaluation, we use a greedy rule selection method to pick the rule at each time step, so only one sequence is generate for each sample. We note that other methods are possible, such as generating multiple sequences and picking the best one. Using the greedy method, it is possible that it generates missing or repeated actions. However, since the cost function is not used during evaluation, we observe that this has minimal impact on the model.

\paragraph{Input Features.} In the experiments, as mentioned, we use VGG, I3D and AssembleNet as initial features.
VGG and I3D use RGB inputs, while AssembleNet (by network design) uses RGB and optical flow as input. The optical flow is computed over RGB inputs on the fly. The CTC and ECTC methods, which are also comprated in the paper, use IDT features \cite{wang2011action} features on the 50-salads and Breakfast datasets and AssembleNet on the NIV dataset, unless otherwise noted. 

\textbf{Specific Model Details}
We provide specific details about the model size. For the various experiments, 
$|\mathcal{H}|$, the size of the set of states, was set to 50 for all experiments and datasets. Changing this value did not significantly impact performance as long as it was greater than the expected number of outputs $|\mathcal{O}|$. $\mathcal{R}$, the set of transition rules, was set to 3 per-state, a total of 150, which is again fixed for all experiments. We use this strategy to be consistent across experiments; this can be further tuned for specific dataset to improve performance.  We set $M=32$, we note that we found the model was not sensitive to this setting, provided it was larger than 8.

\section{Action length equation}
Computing the average action length cost function can be done as:

\begin{equation}
 C_2(S) = \sqrt{\frac{1}{|\mathcal{O}|}\sum_{a\in\mathcal{O}} \left(L(a, S) - \frac{1}{|\mathcal{O}|}\left(\textstyle\sum_{i\in\mathcal{O}}L(i, S)\right)\right) ^ 2},
\end{equation}
where $L(a,S)$ computes the length (i.e., number of frames) labeled as action $a$ in sequence $S$. This function will be minimized when all actions occur for equal number of frames.

\section{Excluded Kinetics classes}
We removed some classes from the Kinetics dataset, used to pretrain the models to obtain the initial features, in order to avoid overlap with the actions we are trying to discover. We also provide a list of some similar classes we left in Kinetics. 

\begin{enumerate}
    \item cooking egg
    \item scrambling eggs
    \item preparing salad
    \item making a sandwich
\end{enumerate}

Similar actions left in:
\begin{enumerate}
    \item peeling apples/potatoes (similar to 50-salads peeling cucumber)
    \item cutting apple/orange/watermelon/pineapple (similar to 50-salads cutting cucumber/tomaoto/cheese)
    \item changing wheel (similar to NIV changing a tire)
    \item planting trees (similar to NIV repotting plant)
    \item frying vegetables (similar to Breakfast frying an egg)
\end{enumerate}

\section{Supplemental Results}
In Table \ref{tab:num-steps}, we report the quantitative results corresponding to Figure 7 in the main paper, for future reference.

\begin{table}[]
\small
    \centering
  
    \label{tab:num-steps}
    \begin{tabular}{l|ccccc}
    \toprule
    \# Steps &  Change Tire (11) & CPR (7) & Repot Plant (7) & Make Coffee (10) & Jump Car (12) \\
    \midrule
    GT Steps & 0.60 & 0.52 & 0.32 & 0.37 & 0.29 \\
    5 &  0.45 &  0.48 &  0.25 & 0.25 & 0.12 \\
    7 &  0.48 & 0.52 &  0.32 &  0.30 & 0.18 \\
    9 &  0.54 & 0.52 & 0.32 & 0.35 & 0.22 \\
    11 & 0.60 & 0.50 & 0.30 & 0.34 & 0.27 \\
    13 & 0.58 & 0.48 & 0.28 & 0.35 & 0.26 \\
    \bottomrule
    \end{tabular}
      \caption{Varying the number of steps used in the model. The number in parenthesis indicates the ground-truth number of steps for each activity. NIV Dataset.}
\end{table}

\end{document}